\documentclass[11pt,a4paper]{article}
\usepackage[hyperref]{acl2020-templates/acl2020}
\usepackage{times}
\usepackage{latexsym}

\usepackage{microtype}
\aclfinalcopy 
\usepackage{xcolor}

\usepackage{graphicx}
\usepackage{subcaption}
\usepackage{multicol}
\usepackage{booktabs}

\title{Detecting anxiety from short clips of free-form speech}

\author{Prabhat Agarwal \\
Stanford University \\
  \texttt{\normalsize prabhat8@stanford.edu} \\\And
   Akshat Jindal\\
Stanford University \\
  \texttt{\normalsize akshatj@stanford.edu} \\\And 
  Shreya Singh \\
  Stanford University \\
  \texttt{\normalsize ssingh16@stanford.edu}} 

\date{2 June 2020}

\title{Detecting anxiety from short clips of free-form speech}

\begin{document}
\maketitle
\begin{abstract}
Barriers to accessing mental health assessments including cost and stigma continues to be an impediment in mental health diagnosis and treatment. Machine learning approaches based on speech samples could help in this direction. In this work, we develop machine learning solutions to diagnose anxiety disorders from audio journals of patients. We work on a novel anxiety dataset (provided through collaboration with Kintsugi Mindful Wellness Inc.) and experiment with several models of varying complexity utilizing audio, text and a combination of multiple modalities. We show that the multi-modal and audio embeddings based approaches achieve good performance in the task achieving an AUC ROC score of 0.68-0.69.
\end{abstract}

\section{Introduction} \label{sec:intro}

Mental and physical health are equally important components of overall health but there are many barriers to accessing mental health assessments including cost and stigma. Even when individuals receive professional care, assessments are intermittent and may be limited partly due to the episodic nature of psychiatric symptoms. Therefore, machine‐learning technology using speech samples obtained in the clinic, remotely over telephone or through recording in personal mobile devices could help improve diagnosis and treatment. In this work, we study the problem of detecting anxiety disorders through speech. Since this is a relatively under-studied problem, the dataset is quite small and hence we plan to explore different approaches to deal with the problem including using hand-crafted features motivated by psychological studies on anxiety and pretrained embeddings models. The dataset provided by Kintsugi provides an interesting opportunity in this area as the recordings are self-recorded in a comforting environment as opposed to a clinical setting. The code is publicly available on Github\footnote{https://github.com/prabhat1081/Anxiety-Detection-from-free-form-audio-journals}.

The paper is outlined a follows: Section \ref{sec:lit} gives details about the existing literature in this problem domain. In Section \ref{sec:data}, we describe the dataset used for training our models along with the task description. In Section \ref{sec:approach}, we provide a detailed explanation of our five models implemented for the task. Section \ref{sec:exp}  outline the evaluation routines and presents the study results. Finally, we conclude our findings in Section \ref{sec:conc}.

\section{Related Works} \label{sec:lit}

Liu et. al. \citep{low2020automated} provides a systematic review of studies using speech for automated assessments across a broader range of psychiatric disorders and summarizes the different datasets, features, and methods that have been employed in this area. The authors summarize the key acoustic features used in these models across disorders. 
Many studies \cite{horwitz2013relative, quatieri2012vocal, cummins2015review, singh2019one} have observed that depression is linked with a decrease in f0 and f0 range and also leads to jitter and shimmer with an increase in depression severity. While jitter and shimmer are also correlated with anxiety ~\cite{ozseven2018voice, silber2016social}, many studies~\cite{gilboa2014being, ozseven2018voice, weeks2012sound, galili2013acoustic} have found a significant increase in mean f0 in patients with an anxiety disorder. Most studies surveyed by the authors employed simple machine learning models like SVM, logistic regression and very few studies used more complex networks with transfer from speech recognition tasks \cite{singh2018footwear}. We plan to utilise these features which have been well studied both in the psychology and machine learning assessment of anxiety and depression in our work and compare them against deep learning models both in terms of performance and generalizability.

State-of-the-art works on detecting mental disorders or emotional states (e.g., anxious vs. calm) from audio data use supervised learning approaches, which must be “trained” from examples of the sound to be detected \cite{kumari2017parallelization}. In general, learning such classifiers requires “strongly labeled”  annotated data, where the segments of audio containing the desired vocal event and the segments not containing that event are clearly indicated. However getting strongly labelled data is tough an in our project too, we only have weakly supervised data. Salekin et. al.~\cite{salekin2018weakly, 10.1007/978-981-10-8639-7_25} presents a weakly-supervised framework for detecting symptomatic individuals from their voice sample. The idea here is to collect long speech audio samples from individuals already diagnosed with or high in symptoms of specific mental disorders from situations that may heighten expression of the symptoms of respective disorders. This type of data is considered “weakly labeled”, meaning that although they provide information about the presence or absence of disorder symptoms, they do not provide additional details such as the precise times in the recording that indicate the disorder etc.

For handling weakly labelled data, they propose a new feature modelling technique called ``NN2Vec''  to generate low-dimensional, continuous, and meaningful representation of speech from long weakly labelled audio data as the other conventional methods work well only with strongly labelled data. To complement that, they present a Multiple instance learning paradigm which is a slight adaptation to classic supervised learning leading to better results. Their NN2Vec and Bi-LSTM MIL approach achieved an F-1 score of 90.1\% and 90\% accuracy in detecting speakers high versus low in social anxiety symptoms. This F-1 scores is 20.7\% higher than of the best baselines they tried.

Recently multi-modal approaches~\cite{siriwardhana2020jointly, al2018detecting} to the speech analysis have shown promising results. Hanai et. al.~\cite{al2018detecting, 10.1145/3308560.3316599} developed an automated depression-detection algorithm that models interviews between an individual and agent and learns from sequences of questions and answers without the need to perform explicit topic modeling of the content \cite{jindal2023classification}. The authors used the data of 142 individuals undergoing depression screening through a human-controlled virtual agent which prompts each individual with
a subset of 170 possible queries that included direct questions (e.g. ‘How are you?’, ‘Do you consider yourself to be an introvert?’), and dialogic feedback ( e.g. ‘I see’, ‘that sounds great’). The data was from the publicly available distress analysis and interview corpus (DAIC) and contains audio and text transcriptions of the spoken interactions \cite{rajan2023shaping, rastogi2023exploring}.

The authors modeled these interactions with audio and text features in Long-Short Term Memory (LSTM) neural network model to detect depression and through various experiments and ablation studies showed that not only did a combination of modalities provide additional discriminative power, but that they contained complementary information and hence it is quite useful to model both.


\begin{figure}[ht]
   \centering
   \begin{subfigure}{0.9\linewidth}
    \centering
    \includegraphics[width=0.9\linewidth]{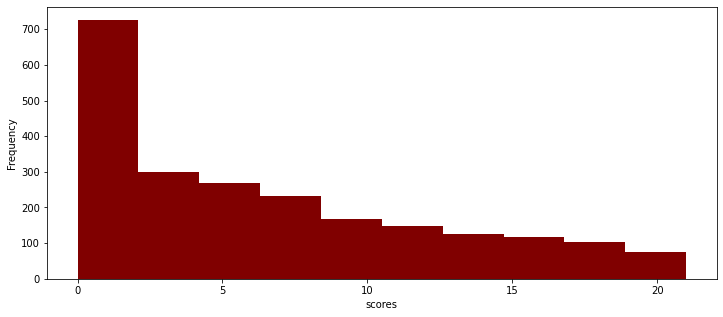}
    \caption{Anxiety score distribution}
    \label{fig:anx_score}
\end{subfigure}
\\
\begin{subfigure}{0.9\linewidth}
    \centering
    \includegraphics[width=0.9\linewidth]{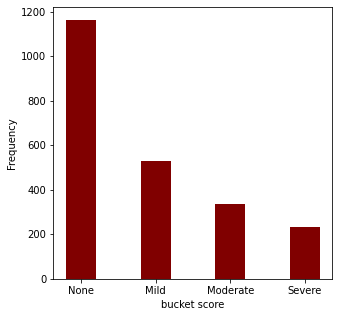}
    \caption{Anxiety level distribution}
    \label{fig:buck_dist}
\end{subfigure}
 \caption{Distribution of raw anxiety score and inferred anxiety levels}
     \label{fig:anxiety-score}
\end{figure}

\section{Dataset and Task Description} \label{sec:data}
\begin{figure}[ht]
    \centering
    \includegraphics[width=0.9\linewidth]{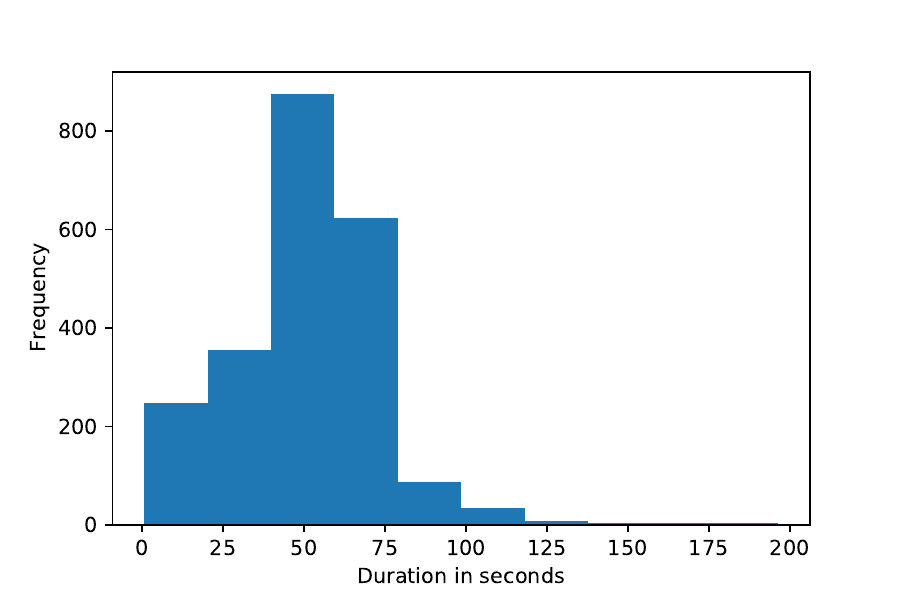}
    \caption{Distribution of the duration of audio files in the dataset}
    \label{fig:dur_dist}
\end{figure}

\begin{figure}[ht]
    \centering
    \includegraphics[width=0.9\linewidth]{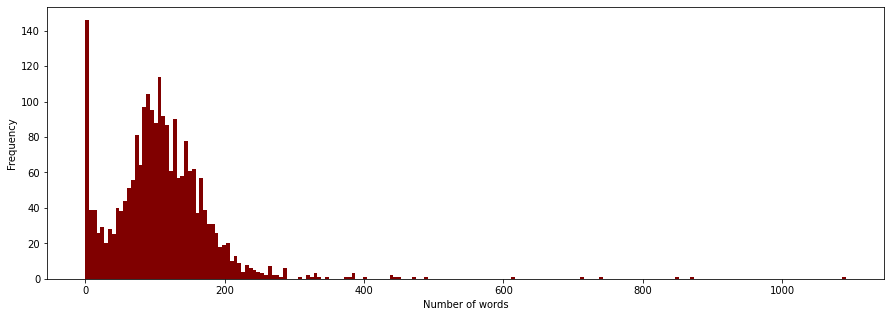}
    \caption{Transcribed words distribution}
    \label{fig:word_dist}
\end{figure}

For our project, we are working with the Generalized Anxiety Disorder (GAD) dataset from Kintsugi Mindful Wellness, Inc. The dataset consists of \textbf{2263} self-recorded audio journals of users along with corresponding score of the user on the GAD-7 questionnaire taken by the user after recording the journal. Each of these files is annotated with an unnormalized anxiety GAD-7 score and a bucketed score. The unnormalized GAD-7 score ranges from 0 to 21 and Fig. \ref{fig:anx_score} shows its frequency distribution. These raw anxiety scores are further split into 4 buckets of anxiety levels, None, mild, moderate and severe, distribution of which is shown in Fig. \ref{fig:buck_dist}. The journals are of varying lengths with most being around 1-2 minutes long. The distribution of the duration of the audio journals in the dataset is shown in Fig.~\ref{fig:dur_dist}. 
The dataset does not contain human transcripts for the journals and hence we use the state-of-the-art Wav2Vec2~\cite{baevski2020wav2vec} to transcribe the given audio files and analyze the textual content. In the transcriptions generated by this model, there are a total of ~244K words and ~18K unique words. The average number of words per conversation is 108. Fig.\ref{fig:word_dist} shows the frequency distribution of the transcribed words.
We further use T5~\cite{raffel2019exploring} base model fine-tuned on emotion recognition and sentiment analysis tasks on the transcribed text using Huggingface~\cite{wolf2019huggingface}. The emotion recognition model classifies each transcribed audio file into one of the 5 emotions, namely, anger, fear, joy, love, sadness while the sentiment analysis model classifies each transcript into two classes, positive and negative.

\begin{figure}[ht]
   \centering
     \begin{subfigure}{0.9\linewidth}
         \centering
         \includegraphics[width=0.8\linewidth]{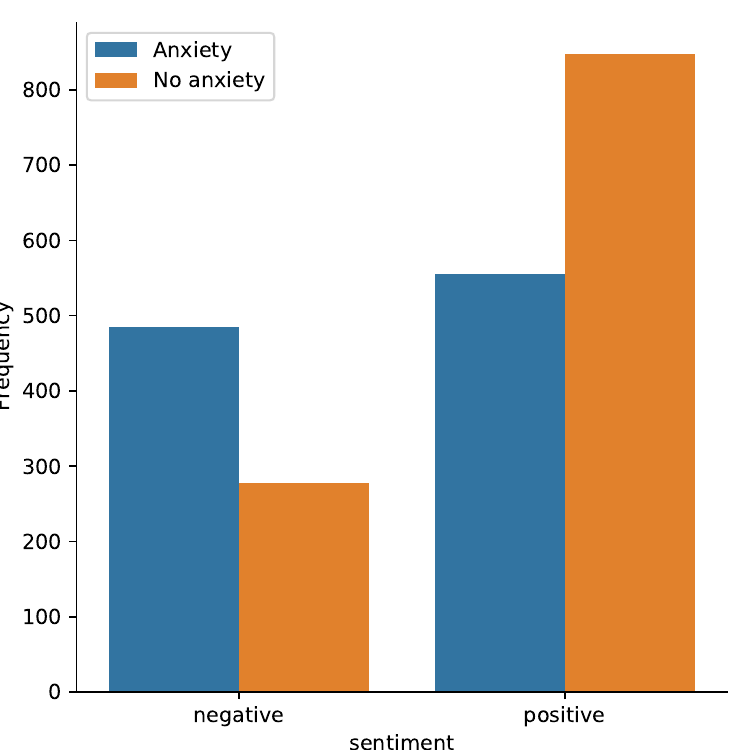}
         \caption{Sentiment distribution in anxiety and non--anxiety classes}
         \label{fig:a}
     \end{subfigure}\\
     \begin{subfigure}{0.9\linewidth}
         \centering
         \includegraphics[width=0.8\linewidth]{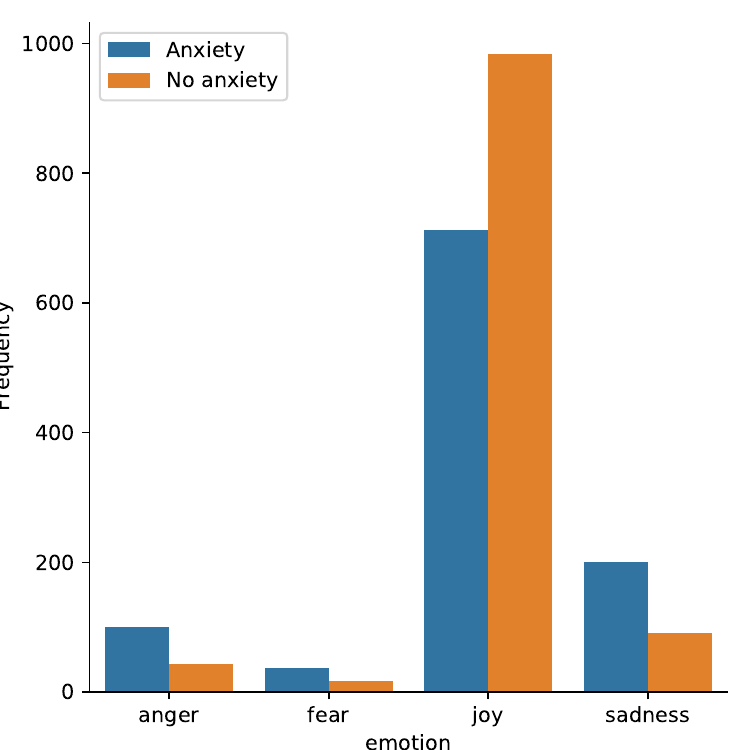}
         \caption{Emotion distribution in anxiety and non--anxiety classes}
         \label{fig:three sin x}
     \end{subfigure}
     \caption{Sentiment and Emotion distribution in anxiety and non--anxiety classes}
     \label{fig:final}
\end{figure}

Fig. \ref{fig:final} shows visualizes the distribution of different sentiments and emotion in transcripts of anxiety and non-anxiety classes. We classify an audio file/datapoint `anxiety' if it belongs to the buckets, mild, moderate or severe, and consequently, `non-anxiety' are those datapoints which fall in the None bucket (from Fig.\ref{fig:buck_dist}). 

\subsection{Task Description}
In this work, we focus on the binary task of detecting whether a user has anxiety disorder or not given their journal recordings. We group the all levels of anxiety, namely mild, moderate and severe as anxiety class for this work. Table \ref{tab:my-table} shows the train, validation and test split statistics of the dataset for the binary classification task (Anxiety vs Non-Anxiety).


\begin{table}[ht]
\centering
\begin{tabular}{@{}lcccc@{}}
\toprule
Class      & Train & Valid & Test & Total \\ \midrule
No anxiety & 840   & 149   & 173  & 1162 \\
Anxiety    & 790   & 139   & 166  & 1095 \\
Total      & 1630  & 288   & 339 & 2257 \\ \bottomrule
\end{tabular}
\caption{Data statistics for different splits}
\label{tab:my-table}
\end{table}

\section{Approach} \label{sec:approach}
In this study, we sought to model self-recorded hournals of users in order to detect if the individual has an anxiety disorder based on GAD-7 score~\cite{spitzer2006brief}. We conducted four sets of experiments with different audio, and text features extracted from the audio data to predict anxiety.
\begin{enumerate}
    \item Model based hand-crafted audio features
    \item Model based on transcript embeddings
    \item Model based on transcript embeddings with sample weights
    \item Model based on low-dimensional audio embeddings using wav2vec
    \item Multi-modal model based both both audio and transcript features
\end{enumerate}

Details of each approach is described below and our code for the experiments can be found online.
\subsection{Hand-crafted audio features}
Work on voice sciences over recent decades has led to a development of acoustic features for different speech analysis tasks. GeMAPS~\cite{eyben2015geneva} standard acoustic feature set for various areas of automatic voice analysis, such as paralinguistic or clinical speech analysis and has proven to be effective in various standard benchmarks. Hence we use the 88 function features described in the GeMAPS extracted using the Opensmile tool~\cite{eyben2010opensmile}. Based on our analysis of the data, we further extend this feature set with 2 features (emotion and sentiment) extracted from the generated transcripts as described in section~\ref{sec:data}. We train a L1-regularized logistic regression model on these 90 features extracted from the data to predict if the individual has anxiety or not. 
\subsection{Transcript embeddings}
In this model, we only use the transcripts generated from the audio and use them for training out anxiety classifier. We don't use any audio features in this baseline and solely rely on the textual features, thereby converting this problem to a natural language domain problem. Specifically, we use Wave2vec2-Base model~\cite{baevski2020wav2vec} to transcribe the given audio data. Upon qualitatively evaluating the generated transcripts, we see that it is a very close transcription with some grammatical errors. Some examples of the transcripts are shown in table~\ref{tab:transcript_sample}. 

\begin{table}[ht]
    \centering
    \begin{tabular}{|p{0.95\linewidth} |}
    \toprule
    I think the best thing that happened for the past week is a being able to get out a for litbit a besides work i think me magrofere mostly been in our house and having enjoyed the outdoors ...\\
    \midrule
I feel eh but lest less and demotivated there's lot of work that i need to do and i folting the ent on my responsibility is at work sometimes i think in my personal life as well is also just general ... \\
\bottomrule
    \end{tabular}
    \caption{Some snippets of generated transcripts from the audio data using wav2vec2}
    \label{tab:transcript_sample}
\end{table}

We embed the generated transcripts in 768 low-dimensional space using the pretrained sentence embedding model Sentence-BERT~\cite{reimers-2019-sentence-bert} which is a RoBERTa model fine--tuned for sentence similarity task. A Gradient Boosting Classifier (GBC) model is trained on these embeddings to classify presence of anxiety. 
\subsection{Transcript embeddings with sample weights}
In our dataset, in addition to having the binary labels of anxiety we also have the raw GAD-7 scores for the users which range between $0$ and $21$. We explore in this model how using these score to assign different weights to different examples might help the a better anxiety prediction model. We utilize a simple linear weighting strategy and define the weights as
$$w_i = \frac{s_i + 1}{22}$$ where $s_i$ is the raw GAD-7 score for the $i^{th}$ datapoint in the dataset and $w_i$ is its corresponding weight. Intuitively, by giving a less sample weight to a `non-anxiety' datapoint and giving a high sample weight to `high-anxiety' datapoint, we are biasing the model to be reduce the number of false negatives at the cost of potentially producing more false positives.

\subsection{Wav2vec audio embeddings}
\begin{figure}[ht]
    \centering
    \includegraphics[width=0.9\linewidth]{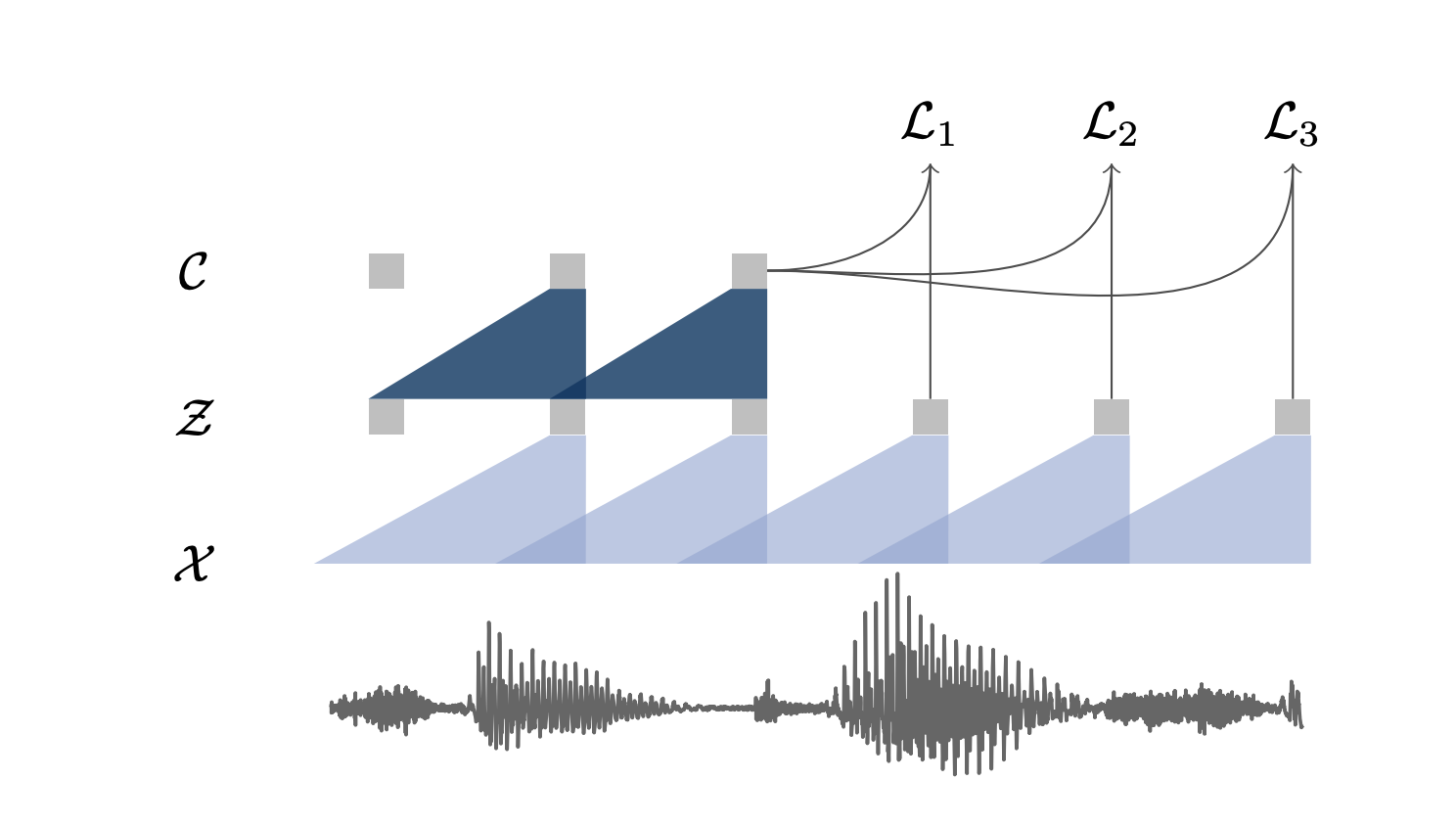}
    \caption{Illustration of the model architecture of wav2vec \cite{DBLP:journals/corr/abs-1904-05862}. We use the mean of the $z$-layer embeddings for our anxiety model.}
    \label{fig:wav2vec}
\end{figure}

Unsupervised representations have shown to be quite effective in various machine learning tasks across different domain like image, text and audio. Hence we explore a model for our task using one of the best unsupervised representation model for speech data. We use Wav2Vec model~\cite{DBLP:journals/corr/abs-1904-05862} to generate 512 dimensional embeddings for the user recordings. The model takes raw audio at sample rate of 16kHz as input and outputs a sequence of 512 dimensional embeddings for the input at two levels, $z$ and $c$ as shown in figure~\ref{fig:wav2vec}. We take the mean of this sequence of embeddings at the $z$ level to construct our final 512 dimensional embedding for the input audio file. We then train a SVM classifier on these embeddings for our anxiety prediction task.

\subsection{Multi-modal model with audio and text}
\begin{figure}[ht]
    \centering
    \includegraphics[width=0.9\linewidth]{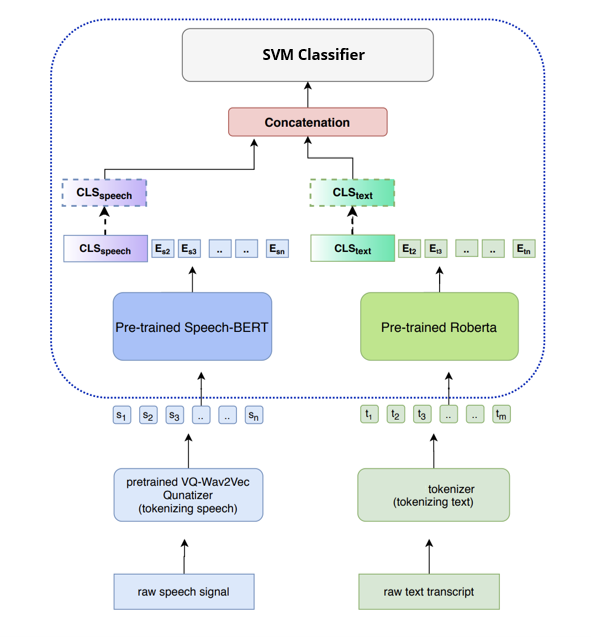}
    \caption{Schematic for Multi-modal anxiety classification model}
    \label{fig:bertlike}
\end{figure}

For our final model, we decided to develop a multi-modal classifcation model combining the information from both the audio and the text transcriptions. Inspired by the multi-modal model for speech emotion recognition in Siriwardhana et. al.~\cite{siriwardhana2020jointly} we generate audio features and text features and use their concatenation to to train a classifier to make predictions for our task. A overall schematic representation of the model is shown in figure~\ref{fig:bertlike}. Details of the different components of the model is described below.
    \begin{itemize}
        \item \textbf{Audio quantizer: VQ-Wav2Vec} \cite{baevski2019vq}: An extension of Wav2Vec~\cite{DBLP:journals/corr/abs-1904-05862}, VQ-Wav2Vec also learns continuous representations by optimizing the CPC objective (Contrastive Predictive Coding) \cite{oord2018representation} and then quantizing it either via a Gumbell Softmax or a K-means approach. For our purposes, we use the pretrained K-means quantizer model.
        \item \textbf{Quantized audio sequence embedder: Speech-RoBERTa} \cite{liu2019roberta}: Then, the quantized audio tokens are fed into a pre-trained RoBERTa model, trained via self-supervision on masked token prediction using quantized vector sequences from Librispeech-960. This gives us dense representations of the audio signal (We only retain the $CLS$ (call it $CLS_{speech}$) token embedding which is the first token embedding) and has been shown to perfrom well for sequence classification tasks.
        \item \textbf{Transcript embedder: RoBERTa} \cite{liu2019roberta}: The text features are generated by tokenizing and feed-forwarding the transcriptions through a standard RoBERTa model pre-trained on the masked LM task as described in \cite{liu2019roberta}. This gives us dense representations of the transcriptions (We only retain the $CLS$ (call it $CLS_{text}$) token embedding which is the first token embedding).
        \item \textbf{SVM Classifier} We concatenate $CLS_{text}$ and $CLS_{speech}$ and then train an SVM classifier on these features to solve the binary classification task.
    \end{itemize}

\section{Experiments} \label{sec:exp}
The results of the experiments, as well as a random baseline are displayed in Table~\ref{tab:baseline-results}.

\begin{table*}[ht]

\centering

\begin{tabular}{@{}lcccc@{}}
\toprule
Model            & Precision & Recall & F1   & AUROC \\ \midrule
Random baseline           & 0.50      & 0.48   & 0.49 & 0.50   \\
Audio features & 0.63     & 0.53   & 0.58 & 0.66  \\ 
Transcipt features & 0.64     & 0.57   & 0.60 & 0.68  \\ 
Transcript features with sample weights & 0.61 & 0.55 & 0.58 & 0.59 \\
Wav2Vec features & 0.66     & 0.61   & 0.64 & 0.69  \\ 
Multi-modal model & 0.66     & 0.60   & 0.61 & 0.68  \\ \bottomrule
\end{tabular}
\caption{Performance of different models}
\label{tab:baseline-results}
\end{table*}

\subsection{Hand-crafted audio features}
\begin{figure}[ht]
    \centering
    \includegraphics[width=0.9\linewidth]{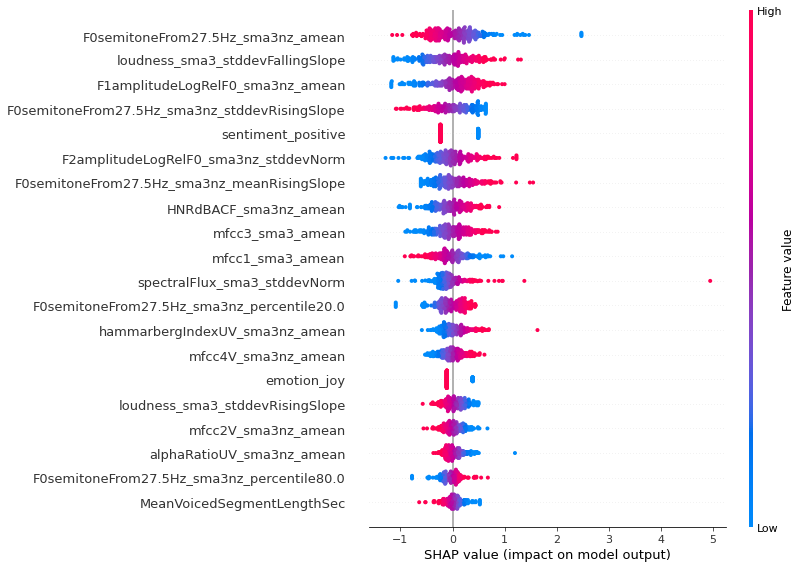}
    \caption{Feature importance of different features for anxiety prediction as obtained using logistic regression model.}
    \label{fig:shap}
\end{figure}

The results of the logistic regression model with L1 regularization is shown in Table~\ref{tab:baseline-results}. The model improves upon the random baseline by around 20\%. We analyse the effect of different features on the model predictions using SHAP~\cite{NIPS2017_7062} and a visualization of the same is shown in Fig.~\ref{fig:shap}. As can be seen from the figure, the top features are loudness, F0, emotion, sentiment, etc. which have been shown in clinical literature to be signals in patients suffering from anxiety. 
\subsection{Transcript embeddings}
The results of the GBC model trained on transcript embeddings are shown in Table~\ref{tab:baseline-results}. The model performs similar to the model based on audio features signifying the importance and effectiveness of text features in mental health related voice classification systems.

The results of adding sample weights to the model is also shown in Table~\ref{tab:baseline-results} and does not improve the performance but rather degrades model performance. Hence we did not experiment with sample weights for our other models.
\subsection{Wav2vec audio embeddings}
\begin{figure}[ht]
   \centering
     \begin{subfigure}{0.9\linewidth}
         \centering
         \includegraphics[width=\linewidth]{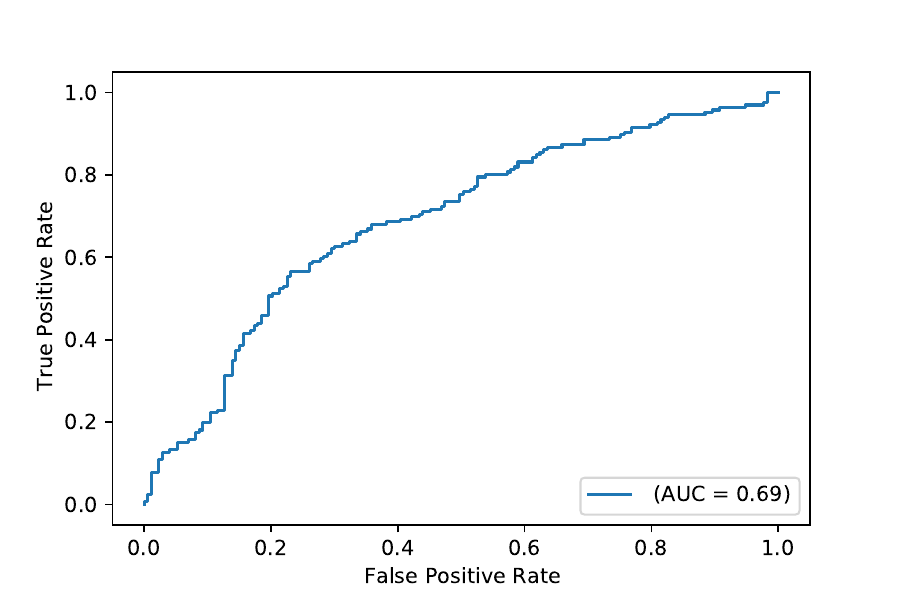}
         \caption{ROC curve}
         \label{fig:roc}
     \end{subfigure}
     \hfill
     \begin{subfigure}{0.9\linewidth}
         \centering
         \includegraphics[width=\linewidth]{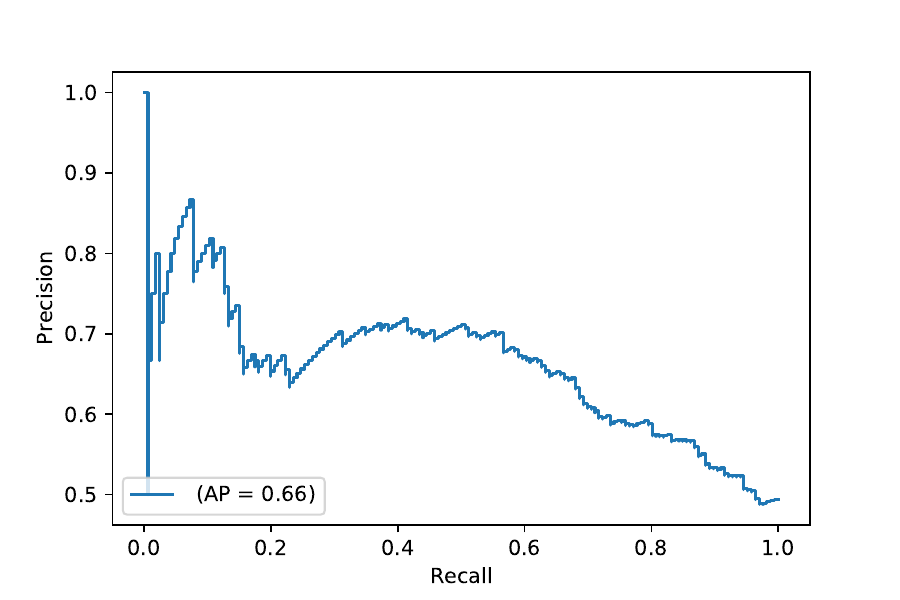}
         \caption{Precision Recall curve}
         \label{fig:prc}
     \end{subfigure}
     \caption{ROC curve and Precision Recall curve for the Wav2Vec features model}
     \label{fig:roc-prc}
\end{figure}

As described in section ~\ref{sec:approach}, we embed the audio recording into 512 low dimensional embedding space and train an SVM classifier to predict anxiety. We tune the different hyperparameters using the validation set and find $C=10$ for perform the best. The results of the best model on the test set are shown in Table \ref{tab:baseline-results}. The ROC curve and the Precision-Recall curve for the model's performance on the test set is shown in figure \ref{fig:roc-prc}. We can see that the model trained on Wav2vec features improved about 30\% over the random baseline. We also analyse the ROC and precision-recall curve for the model and it is shown in Fig.~\ref{fig:roc-prc}. As we can see that the precision is almost the same as recall increases from 0.2 to 0.6 signifying model's robustness.
\subsection{Multi-modal model with audio and text}
For this experiment, as described in Section~\ref{sec:approach}, we firstly used a pre-trained VQ-Wav2Vec model (on Librispeech-960), with a K-means quantizer as provided in the code associated with \cite{siriwardhana2020jointly} to quantize our audio signal. Then we used their pre-trained speech-Roberta model which has the same architecture as the standard RoBERTa-base model \cite{liu2019roberta} to obtain embeddings of size 768 with a max sequence length of 2048. We retain the first token $CLS_{speech}$ embedding. We then create feature representations for text by tokenizing and passing it through a roBERTa-large \cite{liu2019roberta} model to generate embeddings of size 1024 with a max sequence length of 512. We combine the $CLS_{text}$ embedding from to $CLS_{speech}$ to create a 1792 dimensional embedding. We then train an SVM classifier on these features to solve the binary classification task. The results are shown in Table~\ref{tab:baseline-results}. The model does not improve much over the wav2vec embeddings and understanding the reason and improving the model is a direction to explore in the future.

\section{Conclusions} \label{sec:conc}
Mental Health is a very important component of overall health which receives very less attention and importance from the masses. People are more focused towards physical health and ignore their mental well-being which can be disastrous if not dealt with in the earlier stages. In this project, we aim to marry ML-powered speech processing techniques with mental health speech data to automatically detect anxiety from short clips of free-form speech. Our learnings are two fold: (i) Firstly, we implemented various ML powered models of varying complexity on the speech dataset. We start with a random model which gives us about 50\% ROC and treat that as our baseline to improve upon. Our subsequent models are a mix of only text, only audio and a combination of both. (ii) We also gained the experience of working with a very small-sized dataset which can lead to overfitting problems. Hence, our modelling approaches had to be made powerful enough to learn features from a small dataset, while also being robust against overfitting.

In the future, we would like to apply techniques like transfer learning for our task. Our idea is to train our models for depression detection task (which is a larger dataset) and fine-tune it on the anxiety dataset. Due to  similarities between the nature of anxiety and depression disorders, we believe this approach will help us build more powerful models for anxiety detection which are prone to overfitting.
    

\bibliography{acl2020-templates/references}
\bibliographystyle{acl2020-templates/acl_natbib}

\end{document}